\documentclass[10pt,twocolumn,letterpaper]{article}

\usepackage{cvpr}              

\usepackage{graphicx}
\usepackage{amsmath}
\usepackage{amssymb}
\usepackage{booktabs}
\usepackage{graphicx}
\usepackage{url}
\usepackage{multirow}
\usepackage{xcolor}

\newcommand\blfootnote[1]{%
  \begingroup
  \renewcommand\thefootnote{}\footnote{#1}%
  \addtocounter{footnote}{-1}%
  \endgroup
}

\usepackage[pagebackref,breaklinks,colorlinks]{hyperref}

\usepackage[capitalize]{cleveref}
\crefname{section}{Sec.}{Secs.}
\Crefname{section}{Section}{Sections}
\Crefname{table}{Table}{Tables}
\crefname{table}{Tab.}{Tabs.}

\def\cvprPaperID{995} 
\def\confName{CVPR}
\def\confYear{2022}

\begin{document}

\title{Look at here : Utilizing supervision to attend subtle key regions}

\author{Changhwan Lee$^{1 \ast}$ \and Yeesuk Kim$^{1}$ \and Bong Gun Lee$^{1}$ \and Doosup Kim$^{2}$ \and Jongseong Jang$^{3 \dag}$\\
\and
$^{1}$Hanyang University\\
Seoul, Korea
\and
$^{2}$Yonsei University\\
Wonju, Korea
\and
$^{3}$LG AI Research\\
Seoul, Korea
}


\maketitle

\blfootnote{$^{\ast}$ First author {\tt\small lchw711@hanyang.ac.kr}}
\blfootnote{$^{\dag}$ Corresponding author {\tt\small j.jang@lgresearch.ai}}

\begin{abstract}
  Despite the success of deep learning in computer vision, algorithms to recognize subtle and small objects (or regions) is still challenging. For example, recognizing a baseball or a frisbee on a ground scene or a bone fracture in an X-ray image can easily result in overfitting, unless a huge amount of training data is available. To mitigate this problem, we need a way to force a model should identify subtle regions in limited training data. In this paper, we propose a simple but efficient supervised augmentation method called Cut\&Remain. It achieved better performance on various medical image domain (internally sourced- and public dataset) and a natural image domain (MS-COCO$_s$) than other supervised augmentation and the explicit guidance methods.
  In addition, using the class activation map, we identified that the Cut\&Remain methods drive a model to focus on relevant subtle and small regions efficiently. We also show that the performance monotonically increased along the Cut\&Remain ratio, indicating that a model can be improved even though only limited amount of Cut\&Remain is applied for, so that it allows low supervising (annotation) cost for improvement.
\end{abstract}

\section{Introduction}

Deep learning and convolutional neural networks (CNNs) have been successfully applied in various fields, including medicine, visual inspection, and self-driving cars  \cite{hemdan2020covidx,ayan2019diagnosis,chen2020diagnosis,chen2019deep,sa2017intervertebral,yahalomi2019detection,lai2015deep,wang2018interactive,tajbakhsh2020embracing}; however, the utilization of machine learning techniques to recognize a subtle region or a small object in a scene is still challenging \cite{kang2017t,dai2016instance,herath2017going,wu2017image,zhou2019semantic}, especially when sufficient training data are not available on which we frequently encounter the difficulty. Despite the data insufficiency, a model is trained based only on cross-entropy loss to fit the model to the one-hot target label in most classification tasks. As a result, the model is easily overfitted to the training data. To mitigate this problem, regularization methods, including soft labels, temperature scaling, and standard augmentations (e.g., flip, rotation), can be used; however, these are not a kind of explicit methods, which allow the model to recognize subtle and small regions. Occasionally, such as radiological diagnosis and UAV detection\cite{rs13040653}, a model must be trained to look at a relevant region even by utilizing human-guided supervision for ensuring safety. We might consider this is similar to imitation learning\cite{hussein2017imitation} in the field of reinforcement learning.

\begin{figure}
\begin{center}
    \includegraphics[width=1.0\linewidth]{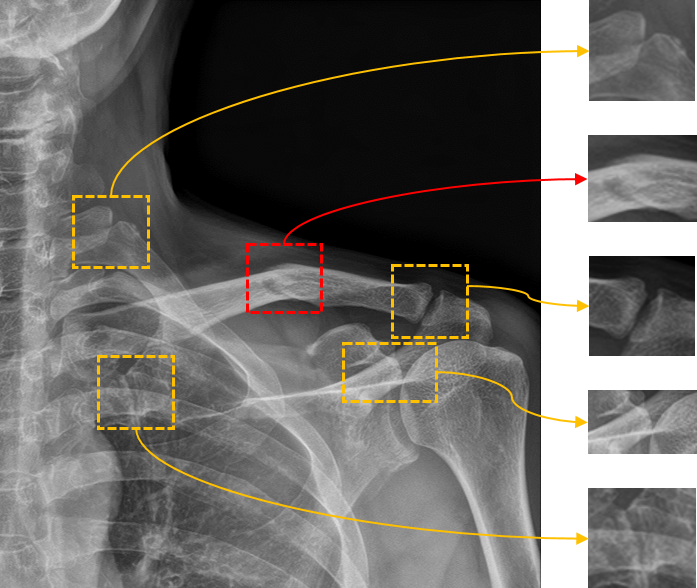}
     \vspace{-0.5cm}
\end{center}
   \caption{Example of candidate features of a lesion. The ground truth and the confusing background region are denoted as red and yellow bounding boxes, respectively.}
\end{figure}

Figure 1 shows a representative X-ray image of a clavicle fracture. A discontinuity is a morphological feature representing a clavicle fracture (red box). However, a lot of discontinuities can be observed in the image (yellow boxes). In contrast to machine learning algorithms, radiologists can correctly diagnose among these discontinuities based on their background knowledge. In the absence of background knowledge, implicit embedding the feature representation of a clavicle fracture is very challenging. It might be solved by considering existing region-perturbed augmentations \cite{krizhevsky2012imagenet,devries2017improved,zhang2017mixup,yun2019Cutmix} or attention-guided methods\cite{yang2019guided,li2018tell} integrated with human supervision.

In this paper, we propose a simple but efficient augmentation strategy, called Cut$\&$Remain, to allow a model to recognize subtle and small key regions. Figure 2 shows the augmented data and its methodology. Relevant area remains and rest area is zero-out, so that the positional information of the lesion and the image size are preserved. 
Especially in radiological diagnosis, the positional information could be a cue for the nature of an area, such as intensity and scale of a lesion. 
Due to the general protocol of medical image acquisition, the anatomical structures are usually aligned so that the target lesion tends to be distributed at a specific position in the image. We show that our method with this information leads to an improvement of the classification performance in the field where the semantics of key features are subtle and amount of data is limited.


The performance of the Cut$\&$Remain in binary (clavicle X-ray dataset), multi-class (pelvic X-ray dataset), and multi-label (Chest X-ray 14 (CXR14)\cite{wang2017chestx} and MS-COCO$_s$, dataset composed of the small objects in MS-COCO 2017\cite{lin2014microsoft}) classification tasks was higher than that of other supervised methods, despite the limited training data and relatively small size of lesions or objects. In addition, the authors qualitatively analyzed the focus region of these models by using Grad-CAM \cite{selvaraju2017grad}, and the results showed that the proposed model effectively focused on the relevant regions. Furthermore, the method was tested based on the number of annotations to investigate the effort required in this task. As a result, its performance monotonically improved with the number of annotations, indicating that, even when limited annotations are available, this method can help the model to learn relevant regions when compared to the baseline.


\section{Related works}
\subsection{Region-perturbed augmentation techniques}
The cutout method \cite{devries2017improved}, in which square regions of the input image are randomly masked out, has been proposed to improve the robustness of CNNs; however, information-rich pixels may be lost during training, which can be critical if the lesion size of the image is relatively small.

Meanwhile, the mixup method  \cite{zhang2017mixup} considers two samples, in which the ground truth label of the new sample is obtained by a combination of one-hot labels (soft label). These samples, however, confuse the model due to their ambiguity and unnaturalness.

The cutmix method  \cite{yun2019Cutmix}, which is a crossover of the cutout and mixup methods, is a novel data augmentation strategy that compensates for the disadvantages of the cutout and mixup strategies. Cutmix produces new samples by cutting and pasting patches within minibatches, thereby enhancing the performance in many computer vision tasks. However, similarly to the cutout strategy, the loss of informative features of the lesion might result in performance degradation.

These methods are useful in combining multiple images or their cropped versions to create new training data; however, they are still not object-aware and have not been designed specifically for small object recognition. Copy-Paste \cite{ghiasi2021simple} is a simple strategy that combines information from multiple images, in an object-aware manner, to copy instances of objects from one image and paste them onto another. However, such an augmentation method for segmentation is difficult to apply in cases where the boundary of the region is anatomically ambiguous, such as fracture and anterior-superior iliac spine. 

\subsection{Learning under privileged information}

Learning using privileged information (LUPI) is a machine learning paradigm where we have provide additional information to network during training that may not be available at test time \cite{vapnik2009new,vapnik2015learning}. This learning paradigm has also been studied for the visual tasks. Unlike max-margin methods, these heavily use the distillation or multitask learning framework.

Hoffman et al.\cite{hoffman2016learning} demonstrated a multi-modal distillation approach to incorporating an additional modality as side information. They first train with a pretrained network and distill the feature information from the privileged network to a main neural network in an end-to-end fashion.

Multi-task learning is a na\"ive way to incorporate privileged information by using auxiliary branch to predict the side information \cite{luowei2019grounded}. The hope is that a the shared feature representation will improve
the target task. However, it does not necessarily satisfy a harmlessness. More importantly, solving the additional task might be more challenging than the original problem.

End-to-end trainable attention mechanisms that allow the network to focus on salient regions for image classification have been studied in the literature \cite{li2018tell,yang2019guided}. Li et al.\cite{li2018tell} proposed attention guided network (AGN) using both classification and attention mining streams to explicitly model the attention mechanism during training. The attention map has been generated from the classification branch using gradient-based Grad-CAM \cite{selvaraju2017grad}.  

In contrast to aforementioned approaches, Cut$\&$Remain requires only the base network.

\begin{figure*}[t!]
\begin{center}
    \includegraphics[width=0.7\linewidth]{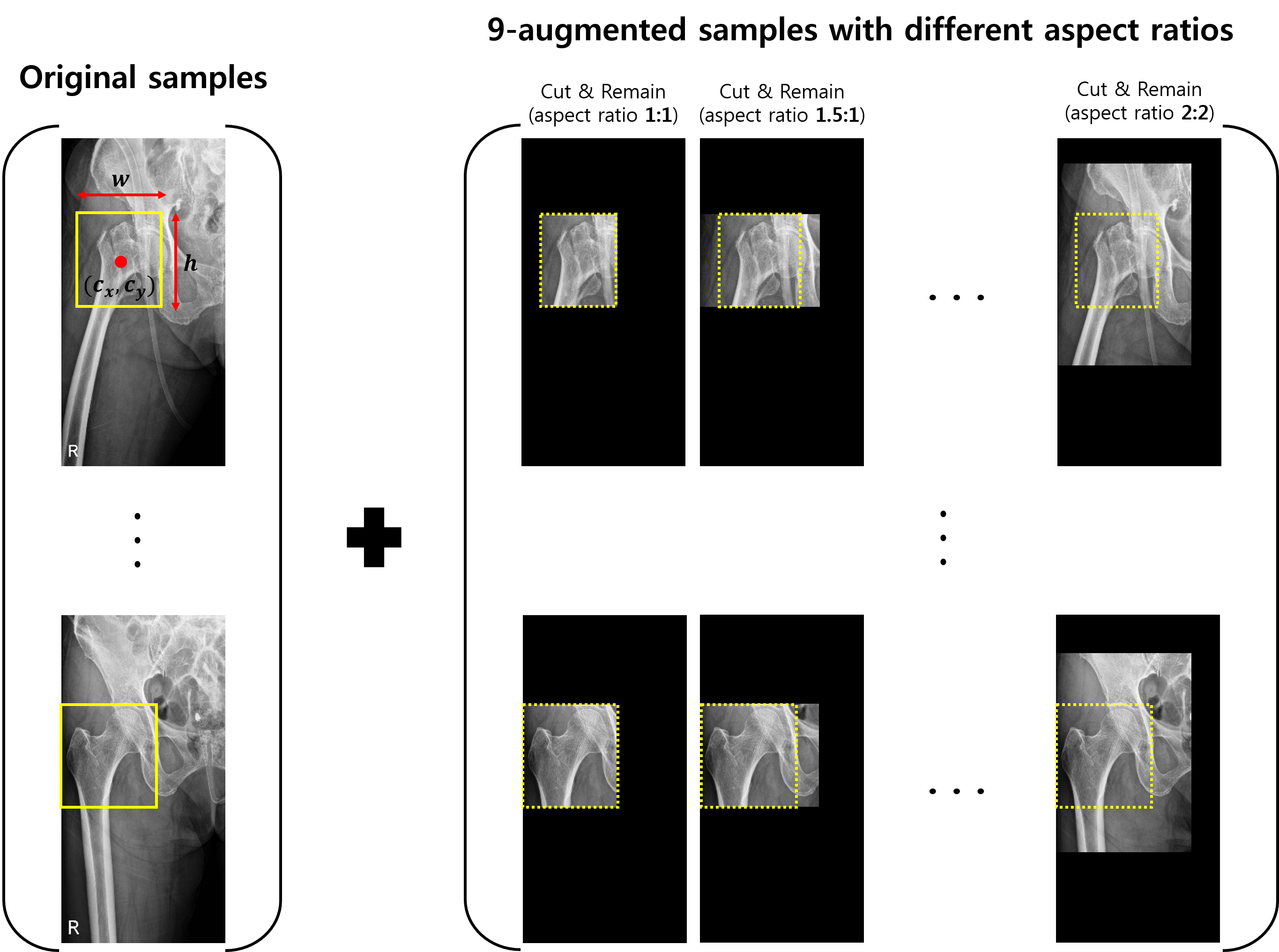}
\end{center}
   \caption{Mini-batch configuration using Cut\&Remain data augmentation during training. The yellow box refers to bounding box annotation $\textbf{B}=(c_{x}, c_{y}, w, h)$. The augmented samples with different aspect ratios are added to current mini-batch(feeding (\textit{original + augmented samples}) to a model during training).}
\end{figure*}

\section{Methods}

Fundamentally, the proposed method utilizes human-annotated bounding box to zero-out unimportant areas in a image while preserving the spatial location of the important areas. This procedure is applied only for a training phase to make a model learn to focus on key regions during training(For sure, it doesn't need any annotation at all in a test phase.) 

Assuming that $x \in \mathbb{R}^{{W}\times{H}\times{C}}$ and $y$ denote a training image and its label, respectively, the goal of Cut$\&$Remain is to generate a new training sample $(\tilde{x},\tilde{y})$,  which can be described by Eq.(1) and used to train the model based on its original loss function.

\begin{equation}
  \begin{array}{l}
    \tilde{x}=\boldsymbol{M}\odot x,\\
    \tilde{y}=y
  \end{array}
\end{equation}

\noindent where $\textbf{M}\in{\{0,1\}}^{{W}\times{H}}$ denotes a binary mask indicating lesion and $\odot$ is element-wise multiplication. To generate Mask $\textbf{M}$, a bounding box annotation, $\textbf{B}=(c_{x}, c_{y}, w, h)$ indicating the region to remain in image $x$ was used, where $c_{x}, c_{y}, w, h$ indicate a box's center coordinate, width and height respectively. To get more augmentations, the aspect ratio of $\{$1.0, 1.5, 2.0$\}$ is applied to box's width and height so that results in nine boxes per single annotation. Then, the element of binary mask $\textbf{M}\in{\{0,1\}}^{{W}\times{H}}$ is set to 1 or 0 according to if it is inside or outside of $\textbf{B}$. 
In this way, Cut\&Remain keeps positional information and scale unchanged for relevant regions, that is distinguished from ``cropping'' that conducts resizing after cutting an area. Please refer to the pseudo code of Cut$\&$Remain in \textit{section 1 in the Supplementary Material}.

In each training step, an augmented sample $(\tilde{x},\tilde{y})$ is generated based on each training sample according to Eq. (1) with different aspect ratios, and included together with original sample $(x, y)$ in the mini-batch, as shown in Figure 2.

\begin{table*}[t!]
\begin{center}
\centering 
\begin{tabular}{lcc} 
\hline 
 & Augmented image $\tilde{x}$ & Augmented label $\tilde{y}$\\ 
\toprule 
Sup-Mixup & $\lambda\textbf{M} \odot x_A + (1-\lambda)\textbf{M}\odot x_B$& $\lambda y_A+(1-\lambda)y_B$\\
Sup-Cutout & Remove random region in $(1-\textbf{M})$ & $y$\\
Sup-Cutmix & $\textbf{M}\odot x_A + (\boldsymbol{1-\textbf{M}})\odot x_B$ & $y_A$ \\

\bottomrule 
\end{tabular}
\end{center}
\vspace{-0.3cm}
\caption{
Supervised-version of augmentation operations. We denote $\boldsymbol{1}$ as a binary mask and $\lambda$ as the combination ratio. $\textbf{M}$ is the binary mask whose value is $\boldsymbol{1}$ inside the annotated box \textbf{B}.} 
\end{table*}

\section{Experiments}
We evaluated the Cut$\&$Remain method considering internally sourced X-ray datasets (clavicle abnormality and femur fracture classification), public Chest X-ray 14 dataset \cite{wang2017chestx}, and the COCO 2017 dataset  \cite{lin2014microsoft} for multi-label classification. For the internal datasets, experienced surgeons annotated the bounding boxes of the relevant areas, including normal images. A digital radiographic examination (CKY Digital Diagnost; Philips, Eindhoven, The Netherlands) included anteroposterior views obtained from patients. 

Conventional region-perturbed augmentations were modified to be a supervised version for fair comparison with Cut$\&$Remain (We also conducted the experiments with original version of them, of course. Please refer to \textit{section 2 in the Supplementary Material}). 
For each training image, we generated a binary mask  \textbf{M}. In each training step, the supervised version of the mixup (Sup-mixup) and cutmix (Sup-cutmix) augmented sample  $(\tilde{x},\tilde{y})$ ) was generated by mixing or combining two randomly selected training samples, $(x_A,y_A)$ and $(x_B,y_B)$, in a mini-batch. A supervised cutout (Sup-cutout) augmented sample was generated by removing a random region in $(1-\textbf{M})$, where $\boldsymbol{1}$ corresponds to a binary mask. The augmentation operations are listed in Table 1, and examples of each augmented image are shown in Figure 3.
 
The experimental results show that Cut$\&$Remain outperforms other data augmentation and attention-guided network (AGN) techniques for all experiments. The AGN to which this method was compared had the same architecture as those proposed in \cite{li2018tell}, including additional supervision. For the COCO 2017 dataset, several categories composed of small objects were selected.

\begin{figure}
\begin{center}
    \includegraphics[width=\linewidth]{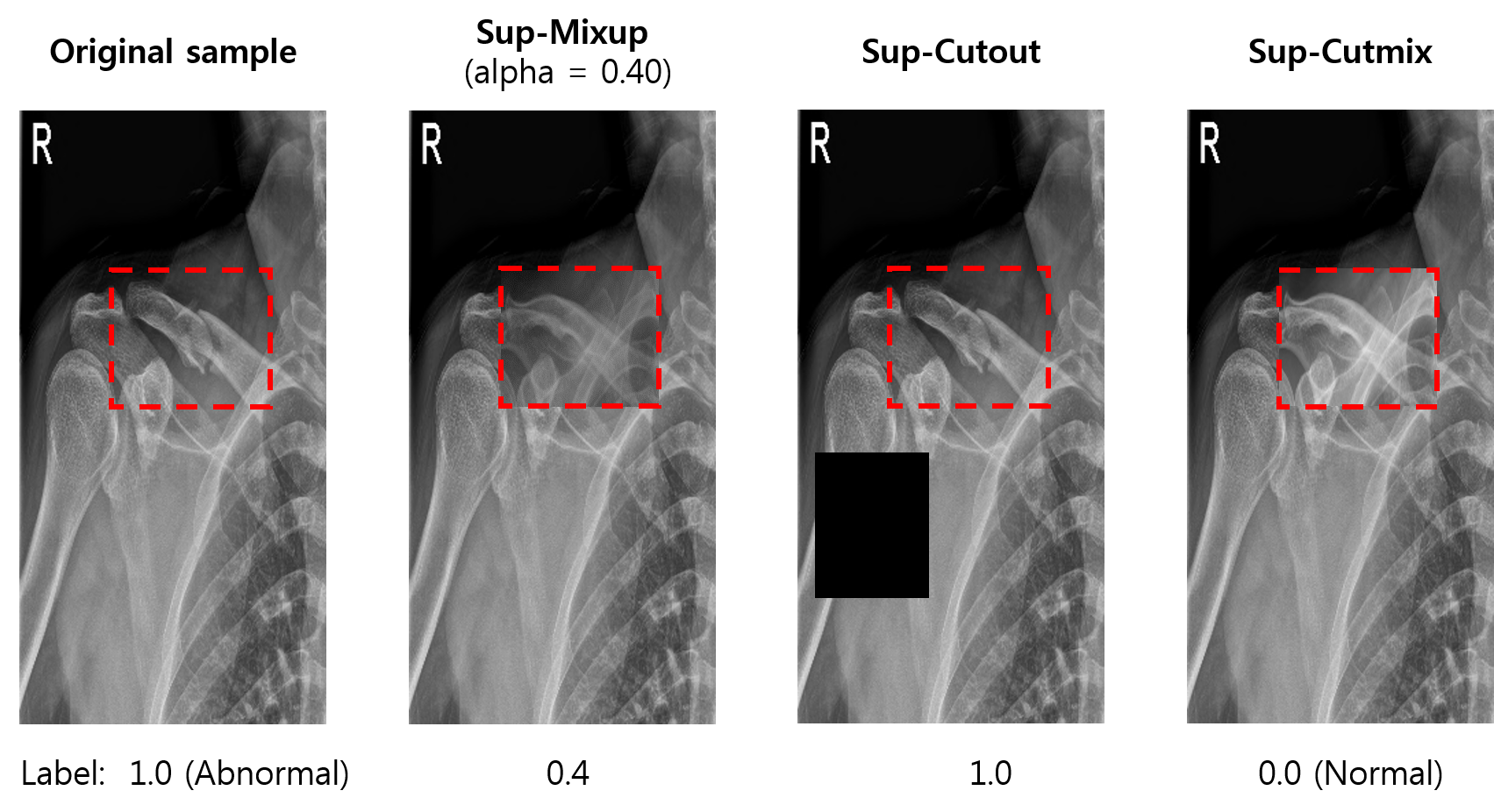}
    \vspace{-0.6cm}
\end{center}
   \caption{Overview of the augmented images from the baseline (Cutout, Mixup, and Cutmix) based on a supervised procedure on the Clavicle X-ray dataset. The red boxes refer to bounding box annotations.}
\end{figure}

\subsection{Binary classification on the clavicle X-ray dataset} 
\paragraph{Experimental setup}
We collected 1,080 clavicle X-rays, including 322 abnormal images (270 fractures and 52 dislocations). The dataset, which was collected from a real-world cohort, also contained cases in which other abnormalities (not only clavicle fracture or dislocation), were present. The overall images were vertically cut in half to increase the number of samples and reduce the resolution of the images. The dataset was split into training (80\%), validation (10\%), and test (10\%) sets. The area under the receiver operating characteristics (AUC-ROC) \cite{fawcett2006introduction} and the F1-score \cite{sasaki2007truth}, obtained after 5-fold cross validation to investigate the classification performance, is reported in this study. We selected ResNet-50 \cite{he2016deep} as the backbone network, and its weights were randomly initialized. Binary cross-entropy loss was used in the classification and the momentum of the employed SGD optimizer was 0.9. The initial learning rate was set to 0.01. The model was trained until 2,000 epochs were performed, and the learning rate was reduced by a factor of 10 at epoch 1,500.


\paragraph{Results}
The experimental results are listed in Table 2. Compared to the supervised version of other augmentation techniques, Cut$\&$Remain showed the highest AUC-ROC (98.6) and F1-score (98.8), and the performance of the base network was improved by more than 7.8 and 7.6, respectively. The results demonstrate that the training of CNNs using random region mixing, removing, or replacing may be affected by irrelevant noisy areas. As a result, bias towards a noisy distractor might be added, harming the generalization performance when the same correlations of the test data are not present in the training data. In contrast, the subtle differences can be exploited in local structures when employing Cut$\&$Remain by highlighting the object of interest and suppressing irrelevant areas. Meanwhile, the utilization of Sup-mixup led to the degradation of the performance of the algorithms in our dataset, as it might provide unnatural artifacts caused by the overlap between tissues. The results of the Cut$\&$Remain support the hypothesis that the lesion must be considered in images that depict a small lesion.

\begin{table}[]
\begin{center}
\centering 
\begin{tabular}{lcc} 
\toprule
    Method & AUC-ROC & F1-score\\
\midrule 
ResNet-50 &  90.8$\pm$\small{1.5} & 91.2$\pm$\small{1.8}\\
Sup-Mixup & 89.0$\pm$\small{1.7} & 90.4$\pm$\small{2.0}\\
Sup-Cutout & 95.6$\pm$\small{1.2} & 95.0$\pm$\small{1.4}\\ 
Sup-Cutmix & 96.8$\pm$\small{1.7} & 97.2$\pm$\small{1.6} \\ 
AGN\cite{li2018tell} & 93.4$\pm$\small{1.1} & 93.0$\pm$\small{1.4}\\ \midrule
Cut$\&$Remain \\ (w/o aspect ratio variation) & 98.4$\pm$\small{0.4} & 98.6$\pm$\small{0.5}\\
Cut$\&$Remain \\ (w/ aspect ratio variation) & \textbf{98.6$\pm$\small{0.4}} & \textbf{98.8$\pm$\small{0.4}}\\
\bottomrule 
\end{tabular}
\end{center}
\vspace{-0.2cm}
\caption{Results of the clavicle X-ray dataset. The averaged AUC-ROC and F1-score and their standard deviation of the 5-fold cross-validation are reported for each set.} 
\end{table}

\begin{table*}[t!]
\begin{center}
\centering 
\begin{tabular}{lcccccc}
\toprule
\multirow{2}{*}{Method}                                                                                                 & \multicolumn{3}{c}{AUC-ROC}                                        & \multicolumn{3}{c}{F1-score}                                       \\ \cline{2-7}
                                                                                                                        & Normal               & A-type               & B-type               & Normal               & A-type               & B-type               \\ \midrule
{ResNet-50}& \small{91.2}\scriptsize{$\pm$0.8} & \small{91.2}\scriptsize{$\pm$1.3}  & \small{87.2}\scriptsize{$\pm$1.8} & \small{92.2}\scriptsize{$\pm$0.9} & \small{73.0}\scriptsize{$\pm$1.1} & \small{47.4}\scriptsize{$\pm$2.5}\\
{\cite{lee2020classification}} & \small{90.6}\scriptsize{$\pm$1.8}                 & \small{91.0}\scriptsize{$\pm$1.0}                 & \small{88.4}\scriptsize{$\pm$1.5}                 & \small{95.4}\scriptsize{$\pm$1.4}                 & \small{88.2}\scriptsize{$\pm$1.8}                 & \small{76.4}\scriptsize{$\pm$2.9}                 \\
{Sup-Mixup} & \small{90.0}\scriptsize{$\pm$1.5}                 & \small{85.8}\scriptsize{$\pm$0.9}                 & \small{86.4}\scriptsize{$\pm$0.7}                 & \small{92.0}\scriptsize{$\pm$1.7}                 & \small{67.2}\scriptsize{$\pm$1.3}                 & \small{63.8}\scriptsize{$\pm$3.9}\\
{Sup-Cutout} & \small{97.0}\scriptsize{$\pm$1.3}                 & \small{96.4}\scriptsize{$\pm$1.5}                 & \small{93.0}\scriptsize{$\pm$1.8}                 & \small{96.4}\scriptsize{$\pm$1.1}                 & \small{85.0}\scriptsize{$\pm$1.7}                 & \small{68.6}\scriptsize{$\pm$3.6}                 \\
{Sup-Cutmix} & \small{95.8}\scriptsize{$\pm$1.6}                 & \small{94.8}\scriptsize{$\pm$1.3}                 & \small{93.0}\scriptsize{$\pm$1.7}                 & \small{96.0}\scriptsize{$\pm$2.0}                 & \small{83.4}\scriptsize{$\pm$2.2}                 & \small{77.0}\scriptsize{$\pm$3.6}                 \\
{AGN \cite{li2018tell}} & \small{94.6}\scriptsize{$\pm$0.6}                & \small{92.8}\scriptsize{$\pm$0.8}                 & \small{90.2}\scriptsize{$\pm$0.8}                  & \small{94.8}\scriptsize{$\pm$1.3}                 & \small{84.4}\scriptsize{$\pm$3.5}                 & \small{66.2}\scriptsize{$\pm$3.4}                 \\   \midrule
{\begin{tabular}[c]{@{}c@{}}Cut\&Remain \\ \footnotesize{(w/o aspect ratio variation)}\end{tabular}}  & \small{97.4}\scriptsize{$\pm$1.3}                 & \small{97.0}\scriptsize{$\pm$1.4}         & \small{97.0}\scriptsize{$\pm$1.1}                 & \small{98.4}\scriptsize{$\pm$1.2}                 & \small{93.0}\scriptsize{$\pm$1.4}        & \small{85.8}\scriptsize{$\pm$2.7}                 \\
{\begin{tabular}[c]{@{}c@{}}Cut\&Remain \\ \footnotesize{(w/ aspect ratio variation)}\end{tabular}}   & \textbf{\small{97.8}\scriptsize{$\pm$0.8}}        & \textbf{\small{97.4}\scriptsize{$\pm$1.1}}                 & \textbf{\small{97.2}\scriptsize{$\pm$1.3}}        & \textbf{\small{98.6}\scriptsize{$\pm$0.8}}        & \textbf{\small{93.8}\scriptsize{$\pm$1.2}}        & \textbf{\small{87.0}\scriptsize{$\pm$2.3}}        \\ \bottomrule
\end{tabular}
\end{center}
\vspace{-0.5cm}
\caption{Results of the pelvic X-ray dataset. The averaged AUC-ROC and F1-score and their standard deviation of the 5-fold cross-validation are reported for each set.} 
\end{table*}

\subsection{multi-class classification of Femur fracture on the pelvic X-ray dataset}
\paragraph{Experimental setup}
A total of 740 anteroposterior pelvic X-ray images were acquired from our institution. Two experienced surgeons reviewed the cases and identified 380 fracture cases and 360 normal cases, following the Arbeitsgemeinschaft Osteosynthese foundation/Orthopedic Trauma Association (AO/OTA) classification standard, in the overall dataset. The dataset was divided into 360 normal (nonfracture), 273 A-type (trochanteric region), and 107 B-type (neck) cases based on the presence of a fracture and its position. The overall images were vertically cut in half to increase the number of samples and reduce the resolution of the images. We split the dataset into 924, 102, and 454 images for training, validation, and testing, respectively, and recorded the AUC-ROC and F1-score for evaluating the classification performance. In all experiments, the ResNet-50 \cite{he2016deep} was selected as our backbone network, and its weights were randomly initialized. We used cross-entropy loss for classification and set the momentum of the employed SGD optimizer to 0.9. The initial learning rate was set to 0.001. The model was trained for 2,000 epochs, and the learning rate dropped by a factor of 10 after epoch 1,500.

\paragraph{Results}
The experimental results are listed in Table 3. On pelvic X-ray images, the AUC-ROC and F1-score obtained after applying Cut$\&$Remain to ResNet-50 were improved by 6.6, and 6.4 for Normal-class classification, respectively.  \cite{lee2020classification} presented a method for classifying femur fractures on X-ray images using deep learning trained with radiology reports. In the literature, they achieved an average F1 score of 81.7 in the 3-class classification task when using the whole images, which is not favorable to translate clinical practice. Conversely, the results of the present study indicate that Cut\&Remain can significantly improve the classification performance due to the high F1 scores and AUCs achieved using this method.

\subsection{Multi-label classification on Chest X-ray 14 dataset}
\paragraph{Experimental setup}
As in \cite{wang2017chestx}, we have examined our model over NIH Chest X-ray dataset. The NIH Chest X-ray dataset consists of 112,120 frontal-view X-ray images with 14 disease labels (each image can have multi-labels); images can have no label as well. Out of the more than 100K images, the dataset contains only 880 images with bounding box annotations; some images have more than one such box, so there are a total of 984 labelled bounding boxes. The 984 bounding boxes annotations are only given for 8 of the 14 disease types. The remaining 111,240 images have no bounding box annotations but do possess class labels. The images are 1024 × 1024, but we have resized them 448 × 448 for faster processing.

In all experiments, the ResNet-50 \cite{he2016deep} was selected as our backbone network, and its weights were randomly initialized. We used binary cross-entropy loss for classification and set the momentum of the employed SGD optimizer to 0.9. The initial learning rate was set to 0.01. The AUC-ROC which are the same statistical variables reported in the conventional settings for Chest X-ray 14 dataset\cite{wang2017chestx,kim2020learning}.

\paragraph{Results}
Table 4 shows test set performance of Cut\&Remain compared with other region-perturbed augmentation techniques. In setting that utilize limited number of supervision (1.26\% of training dataset), we observed that Sup-Mixup degrades performance, which is also showed in other experiments of medical image domain. However, Cut$\&$Remain attains a relatively good performance improvement of 1.46. The results supports our claim that performance can be favorably improved compared to other augmentation techniques even with a very small amount of supervision.


\begin{table}[]
  \centering 
  \begin{tabular}{lc} 
    \toprule
    Method & AUC-ROC\\
    \midrule
    ResNet-50 &  79.32 \\
    Sup-Mixup & 77.56 \\
    Sup-Cutout & 79.68 \\ 
    Sup-Cutmix & 79.51\\
    \midrule
    Cut$\&$Remain (w/o aspect ratio variation) & 80.43\\
    Cut$\&$Remain (w/ aspect ratio variation) & \textbf{80.78}\\
    \bottomrule 
  \end{tabular}
  \caption{Results of the Chest X-ray 14 dataset. The average AUC scores across the 14 diseases are reported.} 
\end{table}

\subsection{Multi-label classification on the COCO 2017 dataset}
\paragraph{Experimental setup}
In this experiment, for generality, we validated the performance of the Cut\&Remain in the natural image domain. The Microsoft COCO 2017  \cite{lin2014microsoft} is a standard dataset built by Microsoft for object detection, image segmentation, and other applications. The training set was composed of 118,287 images that depicted common objects in scenes. These objects are categorized based on 80 classes, and every image contains an average of 2.9 objects. As the ground-truth label of the test set is not available, we evaluated a validation set of 5,000 images. The dataset includes various types of small objects and complex backgrounds; therefore, it is suitable for small object classification. Only images in which the average area occupied by the object is lower than 2\% of the whole image size were added to the dataset to obtain only the images containing small objects. Based on this criteria, we selected 66,612 images for training and 2,805 images for testing that included 27 types of objects, including cars, traffic lights, kites, and cups, to create a training dataset, which was named MS-COCO$_s$ (see \textit{section 2.4 in the Supplementary Material} for details). The mean average precision (mAP), average per-class F1 (CF1), and average overall F1 (OF1), which are the same statistical variables reported in the conventional settings for COCO 2017 \cite{liu2018multi,wang2020multi}, were calculated. For all experiments, we employed ResNet-50 \cite{he2016deep} as our backbone network, whose weights were randomly initialized. We used binary cross-entropy loss for classification and an Adam optimizer. The initial learning rate was set to 0.001. The model was trained for 50 epochs, and the learning rate dropped by a factor of 10 at epoch 40.    

\paragraph{Results}
Table 5 indicates that Cut\&Remain produced better results than those of the baseline on MS-COCO$_s$. mAP was 2.3\% higher than that of the baseline network. However, the performance improvement using Cut\&Remain was relatively small when compared to the previous tasks, which considered medical images. This is caused by the objects that are distributed in various locations in the natural image domain. 
 
Although originally developed for medical image tasks where positional consistency is existed, the classification performance in the natural image domain was also improved by Cut\&Remain. This implies Cut\&Remain can contribute to distinguish a subtleness of small objects in the natural images where positional consistency is not guaranteed, as well as small lesions in the medical images.
The class-wise precision on the MS-COCO$_s$ is shown in Figure 4.

\begin{table}[]
\begin{center}
\centering 
\begin{tabular}{lccc} 
\hline
    Method & mAP & CF1 & OF1 \\
\toprule
ResNet-50 &  74.4 & 69.3 & 74.0\\
Sup-Mixup & 73.5 & 68.6 & 73.0\\
Sup-Cutout & 75.2 & 69.1 & 74.0\\
Sup-Cutmix & 75.6 & 71.1 & 75.0\\ 
AGN\cite{li2018tell} & 76.2 & 71.0 & \textbf{75.7}\\ \midrule
Cut$\&$Remain \\ (w/o aspect ratio variation) & 75.8 & 70.8 & 75.1\\
Cut$\&$Remain \\ (w/ aspect ratio variation) & \textbf{76.7} & \textbf{71.8} & 75.6\\
\bottomrule 
\end{tabular}
\end{center}
\vspace{-0.2cm}
\caption{Multi-label classification results on the MS-COCO$_s$ testset. All metrics are in \%. Results are reported for an input resolution of 448.}  
\end{table}

\begin{figure}
\begin{center}
    \includegraphics[width=\linewidth]{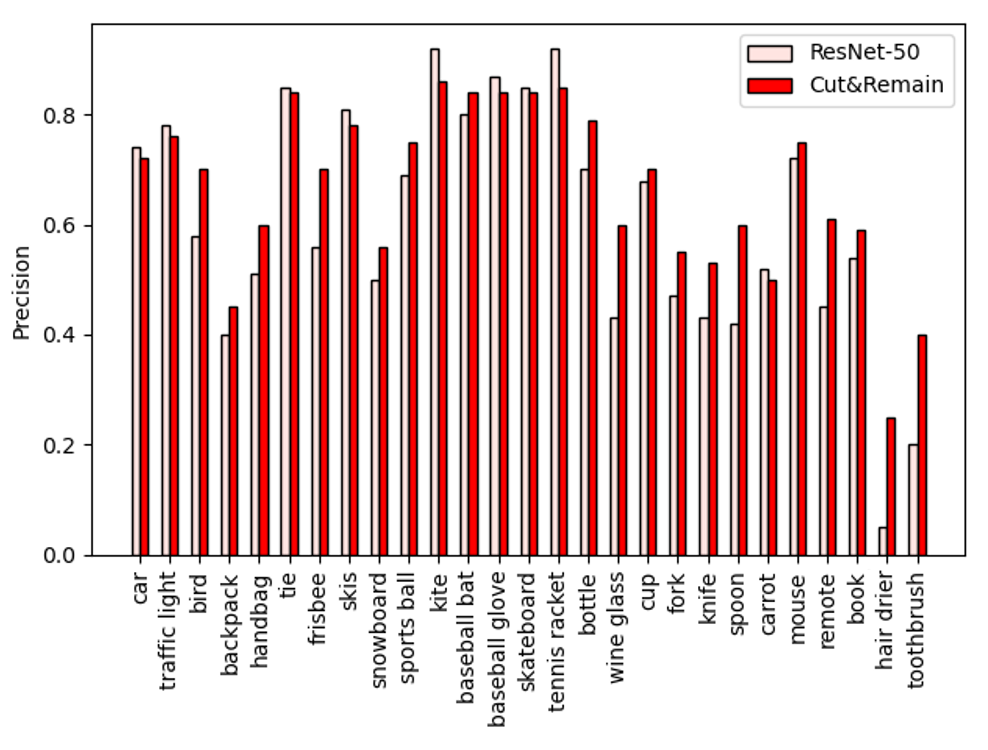}
    \vspace{-0.5cm}
\end{center}
   \caption{The class-wise precision on the MS-COCO$_s$ dataset. The bars represent the results achieved by using baseline and Cut$\&$Remain, respectively. Cut$\&$Remain leads even better results at difficult classes, such as hair drier, backpack, wine glass, and toothbrush}
\end{figure}

\begin{figure*}[t!]
\begin{center}
    \includegraphics[width=\linewidth]{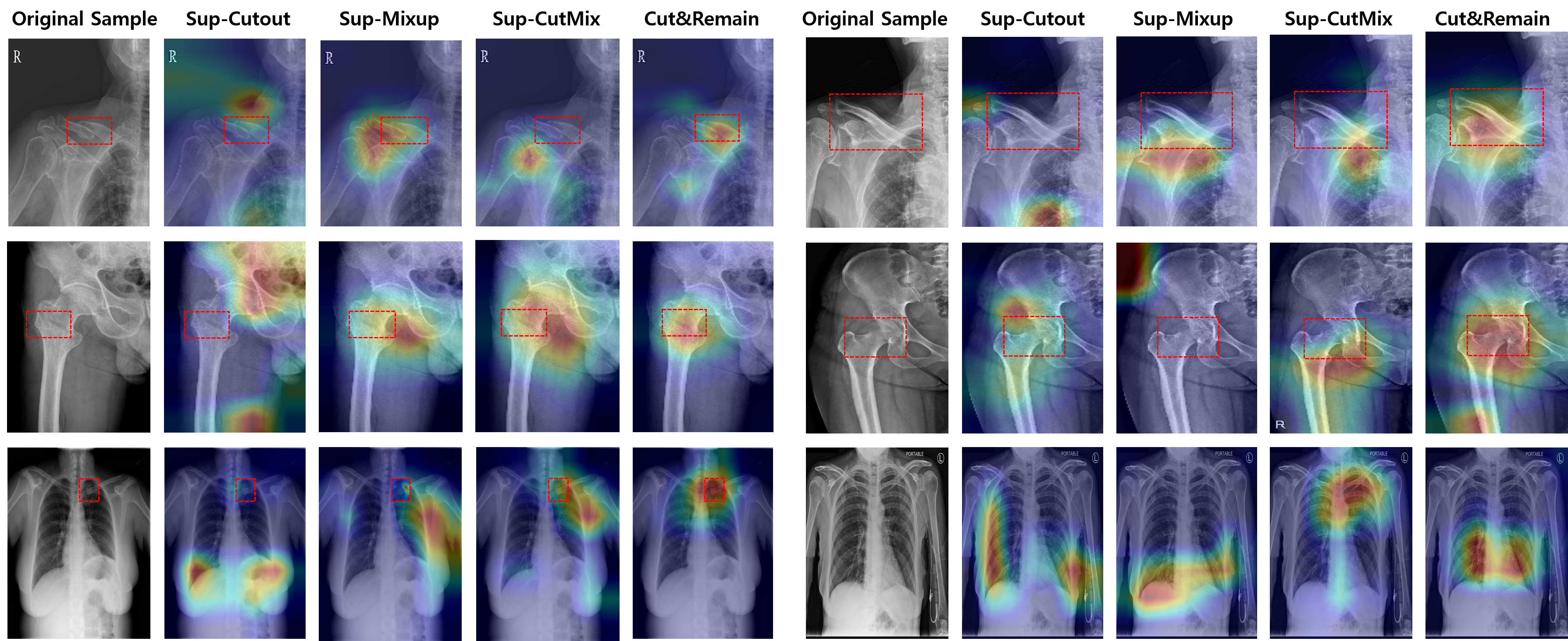}
    \vspace{-0.5cm}
\end{center}
   \caption{Grad-CAM visualization on test images using the model trained with each augmentation technique. Left: Grad-CAM for abnormal class. Right: Grad-CAM for normal class. Ground-truth annotation are shown as a red boxes. }
\end{figure*}

\begin{figure*}[t!]
\begin{center}
    \includegraphics[width=1.0\linewidth]{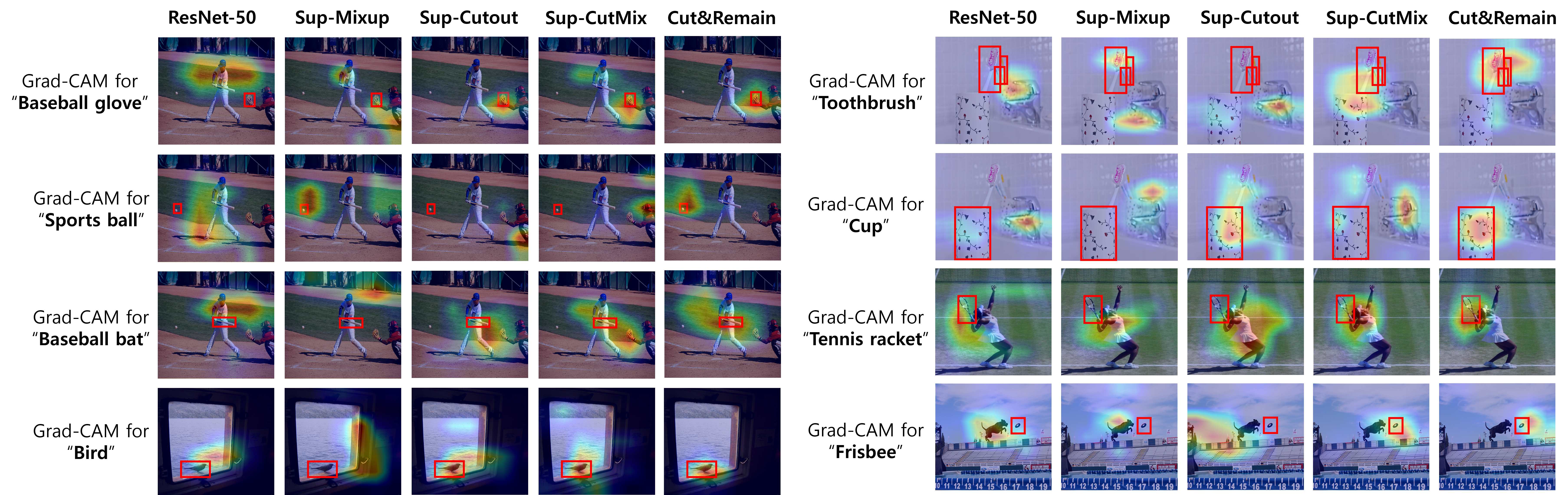}
    \vspace{-0.5cm}
\end{center}
   \caption{Grad-CAM visualization on MS-COCO$_s$ test images. Ground-truth annotation are shown as a red boxes.}
\end{figure*}

\subsection{What does a model learn with Cut$\&$Remain?}
\subsubsection{Qualitative evaluation by Grad-CAM}
We have validated Cut\&Remain such that lesions are mainly considered as cues for classification and the motivation shared by attention-guided networks. To verify that Cut\&Remain recognizing the key lesion on target images after the learning procedure, the activation maps of the test images trained by Cut\&Remain were compared to those trained with the Sup-Cutout, Sup-Mixup, and Sup-Cutmix. Figure 5 shows the test examples and the corresponding Grad-CAM for the abnormal classes on medical image domain. We used the vanilla ResNet-50 model to obtain the Grad-CAM and observe the effect of only using the augmentation method. For the images in which no abnormalities were present, Cut\&Remain could also learn the corresponding mask, thus distinguishing normal from other abnormal classes.

We observed that Cut$\&$Remain allowed a model to detect the lesions. In contrast, mixing introduces unnatural artifacts; therefore, the corresponding model was confused when choosing cues for recognition, as shown in the Grad-CAM, which might lead to suboptimal classification performance, as presented in Tables 2, 3 and 4.

We also visualize the attended regions in the MS-COCO$_s$ dataset, as shown in Figure 6. Cut$\&$Remain could also learn the corresponding context in the natural image domain. Examples containing more samples are presented in the supplementary material.

\begin{table*}[t!]
  \centering 
  \begin{tabular}{lcccc} 
    \toprule
    & Sup-Cutout & Sup-Mixup & Sup-Cutmix & Cut$\&$Remain \\
    \midrule
    Euclidean distance &  35.60$\pm$16.34  & 40.56$\pm$25.12  & 37.53$\pm$17.82  & \textbf{27.45$\pm$13.31}  \\
    Cosine distance &  0.73$\pm$0.16  & 0.77$\pm$0.18  & 0.79$\pm$0.17  & \textbf{0.63$\pm$0.16}  \\
    \bottomrule
  \end{tabular}
  \caption{Average feature vector similarity between the original and augmented samples by a supervised version of the cutout, mixup, cutmix, and Cut$\&$Remain. Each vector was obtained from the last convolution layer of the ResNet-50 trained with the clavicle dataset} 
\end{table*}

\subsubsection{Feature representation property}
To verify that Cut\&Remain assists in generating a background-independent feature vector (i.e., whether the model mainly focuses on the lesion), we analyzed the similarity of the vectors produced by using the original and augmented images. We compared the ResNet-50 trained without any augmentation techniques and individually trained using the Sup-Cutout, Sup-Mixup, Sup-Cutmix, and Cut\&Remain strategies. We conducted this investigation based on the training set of clavicle X-rays and the same experimental setting as the subsection of \textit{``4.1 Binary classification on the clavicle X-ray dataset''}. The Euclidean distance and cosine distance were the similarity measures calculated considering the feature vectors obtained from the output of the last convolution layer.

The experimental results are presented in Table 5. We observed that Cut$\&$Remain created similar representation vectors for the original and augmented samples. In contrast, conventional augmentation techniques, which randomly remove, mix, or replace regions in images, resulted in increased dissimilarity due to the informative features that might have been lost.

\vspace{-0.2cm}

\subsubsection{Performance according the number of annotations  }

Because Cut$\&$Remain utilizes annotations by human effort, this might be a limitation for scalability. It leads the experiment to figure out the relationship between the performance and the amount of annotations. So, we evaluated the performance according amounts of the training dataset applying Cut\&Remain with the clavicle X-ray dataset. The ratio $\gamma$ of the training data applying Cut\&Remain was \{0, 0.2, 0.4, 0.6, 0.8, 1.0\}. The performance of Cut\&Remain with different $\gamma$ is given in Figure 6. The Cut\&Remain achieved the best performance after augmentation was applied to the overall training dataset (i.e., $\gamma=1.0$). Furthermore, the AUC-ROC and F1-score monotonically increased with $\gamma$\ ; therefore, the performance improvement is guaranteed even if limited number of annotations are available. For the experiment on MS-COCO$_{s}$, please see the supplementary.

\begin{figure}
\begin{center}
    \includegraphics[width=1.0\linewidth]{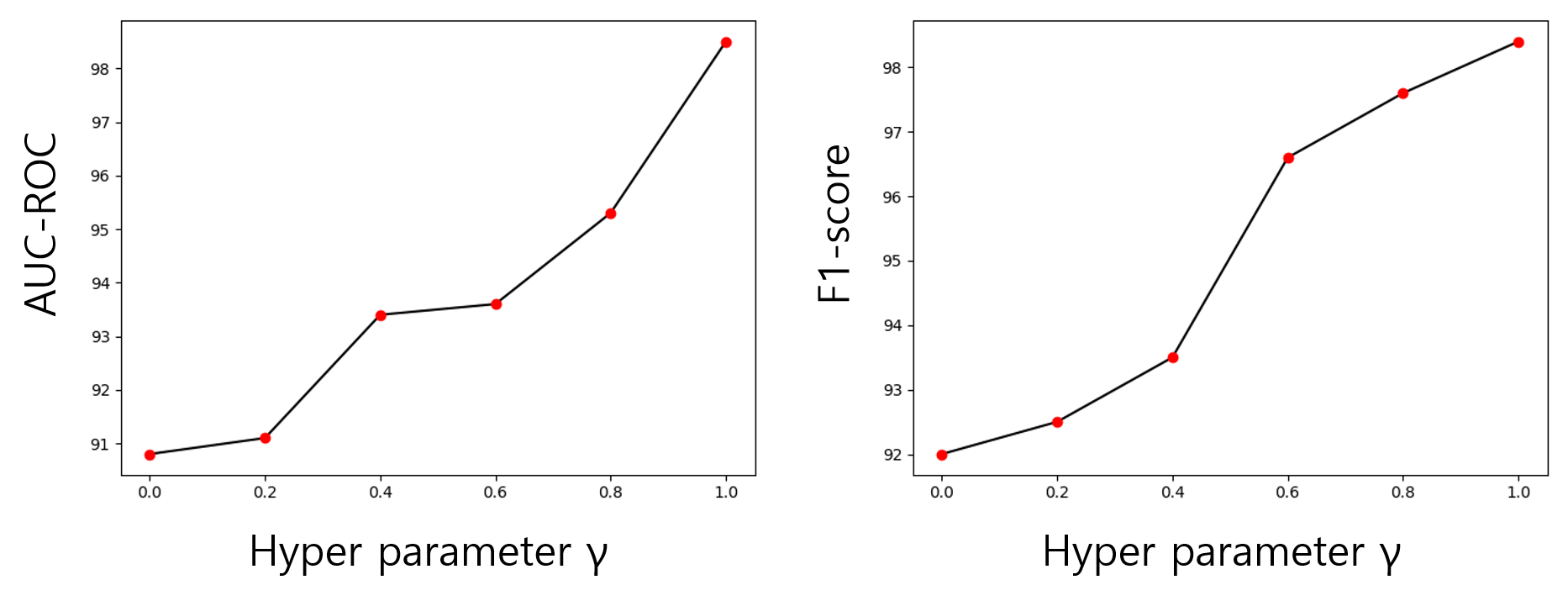}
     \vspace{-0.8cm}
\end{center}
   \caption{Effect of the Cut$\&$Remain and augmented images in the training dataset (in Clavicle X-ray dataset)}
\end{figure}

\section{Conclusion}

We introduced a novel data augmentation strategy, namely Cut$\&$Remain, for training CNNs. This strategy has a strong attention-guided benefit, can be easily implemented, and has no computational overhead. Furthermore, it is surprisingly effective on medical image datasets. We have shown that Cut$\&$Remain can significantly improve the performance of the CNN classifier on various image domains, despite the limited amount of training data and relatively small lesion size. In particular, on clavicle X-ray classification, the AUC-ROC and F1-score of the ResNet-50 were improved by 7.8 and 7.6, respectively, when the Cut$\&$Remain method was applied. On femur fracture classification, Cut$\&$Remain resulted in an AUC-ROC improvement of 6.6 for Normal-class classification when compared to that of the baseline. The explicit attention mechanism, however, did not guarantee performance improvement. On Chest X-ray 14 classification, Cut$\&$Remain resulted in an AUC-ROC improvement of 1.46 compared to the baseline, despite the use of limited number of supervision.

In the natural image domain, MS-COCO$_s$, this data augmentation technique provided consistent improvements over the baseline and other augmentation techniques. Moreover, Grad-CAM analysis and t-SNE visualization in the Supplementary material indicated that Cut$\&$Remain resulted in more focus on lesions, irrespective of the background.

\paragraph{Limitation}Cut$\&$Remain should have been proved with more complex dataset that large and small objects exist together for scalability. It needs to be investigated if an annotation of a large object is necessary or that of small one is enough. 

\newpage

{\small
\bibliographystyle{ieee_fullname}
\bibliography{egbib}
}

\end{document}